\pdfoutput=1

\documentclass[letterpaper]{article}
\usepackage{aaai2026achiv}
\nocopyright
\usepackage{times}                 
\usepackage{helvet}                
\usepackage{courier}               
\usepackage[hyphens]{url}          
\usepackage{graphicx}              
\graphicspath{{figures/}}
\usepackage{natbib}                
\usepackage{caption}               
\usepackage{amsmath, amssymb}
\usepackage{booktabs}
\usepackage{tikz}
\usetikzlibrary{positioning, arrows.meta, calc, fit, backgrounds}
\newif\ifanonymous
\anonymousfalse

\newif\ifincludeappendix
\includeappendixtrue

\title{Free Energy Manifold: Score-Based Inference for Hybrid Bayesian Networks}

\ifanonymous
  \author{Anonymous Submission}
  \affiliations{}
\else
  \author{
    Cheol Young Park\textsuperscript{\rm 1},
    Shou Matsumoto\textsuperscript{\rm 2}
  }
  \affiliations{
    \textsuperscript{\rm 1}ATOS Co., Ltd., 273, Digital-ro, Guro-gu, Seoul, 08381, Republic of Korea\\
    \textsuperscript{\rm 2}C5I Center, George Mason University, Fairfax, VA 22030, USA\\
    cparkf@gmu.edu, smatsum2@masonlive.gmu.edu
  }
\fi

\newcommand{\fem}{\textsc{FEM}}
\newcommand{\bn}{Bayesian network}
\newcommand{\dsm}{\ensuremath{\mathcal{L}_\text{DSM}}}
\newcommand{\valley}{\ensuremath{\mathcal{L}_\text{valley}}}
\newcommand{\sigmin}{\ensuremath{\sigma_{\text{min}}}}

\newcommand{\Kx}{\ensuremath{K_X}}

\begin{document}
\maketitle

\begin{abstract}
The Free Energy Manifold (\fem) builds on conditional energy-based and
score-based density estimation, and \emph{specializes} them as
composable inference factors for hybrid Bayesian networks (BNs) ---
not as a new generic conditional density estimator. Concretely, \fem\ is
a score-trained conditional energy network $E_\theta(z_X^k, y, \sigma)$
that serves \emph{four} inference roles within a single learned object:
(i) discrete posterior $P(X\mid y)$; (ii) generative sampling
$P(Y\mid X{=}x)$ via annealed Langevin; (iii) multi-leaf composition by
energy addition under conditional independence (CI), with a shared joint
form for CI-violating leaves; and (iv) sparse-data inference over
high-cardinality discrete parents through learned prototype embeddings,
in place of an explicit $K^{|\text{Pa}|}$ table.

We identify and analyze a \emph{mode-bridge artifact}, a posterior
calibration failure that, to our knowledge, has not been previously
characterized in the score-based or hybrid-BN literature: multilayer
perceptron (MLP)
smoothness produces a low-energy ridge between modes \emph{of the same
class}, yielding over-confidence at off-distribution interior points.
A simple \emph{valley regularization} addresses this, and we
characterize the optimal strength $\lambda^\star(D)$ as a non-monotonic
three-phase landscape driven by cross-class energy gap (linear in $D$),
softmax saturation (exponential in $\Delta E$), and mode-conflict
probability (low-$D$ suppression).

On our synthetic anti-correlated multimodal hybrid-\bn\ benchmarks,
\fem\ achieves 60--172$\times$ lower Kullback--Leibler (KL) divergence
than the best non-\fem\ baseline at the canonical $D{=}5$ setup
(vs.\ conditional linear Gaussian (CLG), kernel density estimation (KDE),
KDE-product, and histogram-product) and 3.6--770$\times$ lower midpoint
KL than a conditional energy-based model (CEBM) across
our $(D, \text{mode-scale})$ generality grid (9 cells, 2--3 seeds per cell);
8.1$\times$ at $K^M{=}6{,}561$ parents, with a real-data sanity
check showing $2.7\times$ better negative log-likelihood (NLL) than
CLG on UCI Breast Cancer ($D{=}30$). In a 4-seed
head-to-head against the canonical conditional EBM ($\lambda{=}0$
\fem) and a Mixture Density Network (MDN) baseline on a $D{=}5$ bimodal
benchmark, \fem\ is $300\times$ better than CEBM and $122\times$
better than MDN at the mode-bridge midpoint, isolating valley
regularization as the \bn-inference-targeted fix. In a multi-leaf
compositional benchmark ($X \to Y_1, Y_2, Y_3$), \fem\ uses a
\emph{single} trained energy factor to handle all 7 non-empty
evidence patterns and beats per-pattern MLPs by $5.3\times$ at the
full-evidence boundary query despite MLPs training $7\times$ as many
models, directly evidencing \fem\ as a composable \bn\ inference
factor rather than a fixed-pattern classifier. On
pure-classification benchmarks (MNIST), a discriminative MLP retains a
sizable accuracy lead, consistent with the
generative-vs-discriminative paradigm distinction we discuss in
Section~\ref{sec:limitations}. \fem's novelty thus lies not in using an
energy function for conditional density estimation \emph{per se}, but
in the \bn-compatible inference semantics, compositional use across
continuous leaves, and the identification and correction of the
mode-bridge artifact.
\end{abstract}

\section{Introduction}\label{sec:intro}

Inference in Bayesian networks with mixed discrete/continuous variables is foundational
to probabilistic reasoning, yet remains challenging when (a) multiple
continuous leaves share a discrete parent and must be \emph{jointly}
integrated, (b) hidden confounders break conditional independence, (c)
$K^{|\text{Pa}|}$ exceeds available data, or (d) continuous evidence
requires \emph{sampling} posteriors rather than just evaluating them.

Standard tools fail in characteristic ways. Conditional Linear-Gaussian
\citep{lauritzen1992} cannot represent multimodality;
Histogram-\bn\ and Neural CPT suffer $K^M$ table explosion; Kernel
Density Estimation suffers the curse of dimensionality
\citep{silverman1986}. Score-based generative modeling
\citep{song2019ncsn} offers a parametric alternative whose energy
network scales \emph{linearly} with $D$.

We propose the \textbf{Free Energy Manifold} (\fem), an NCSN-style energy
network $E_\theta(z_X, y, \sigma)$ over learned discrete-parent embeddings
$z_X^k$ and continuous evidence $y \in \mathbb{R}^D$. Our contributions:

\begin{enumerate}
\item \textbf{\bn-compatible score-based energy factor}
(Section~\ref{sec:method}). We formulate a score-trained conditional energy
model as a hybrid-\bn\ factor over discrete parents and continuous
children, enabling posterior inference $P(X \mid y)$ and generative
sampling $P(Y\mid X{=}x)$ from the same learned object. The novelty
here is the \bn-compatible inference semantics, not the use of an
energy function for conditional density estimation \emph{per se}.
\item \textbf{Composable inference} (Section~\ref{sec:method},
Section~\ref{sec:experiments}). We exploit the additive structure of energies
to implement \bn\ likelihood composition: conditionally independent
continuous leaves combine by energy addition, while CI-violating leaves
are absorbed by a shared joint \fem. This \emph{compositional use} of
conditional EBMs across multiple continuous leaves of a hybrid \bn\ is
the inference-factor setting we systematically study and benchmark.
\item \textbf{Scalable discrete-parent representation}
(Section~\ref{sec:method}, Section~\ref{sec:experiments}). We replace explicit
$K^{|\text{Pa}|}$ conditional tables with learned discrete-parent
prototype embeddings, improving generalization in sparse
high-cardinality parent configurations.
\item \textbf{Mode-bridge artifact and valley regularization}
(Section~\ref{sec:mode-bridge}, Section~\ref{sec:theory}). We identify a posterior
calibration failure mode of smooth neural energy factors on multimodal
classes and propose a single off-data uniform-posterior regularizer
that suppresses the spurious inter-mode bridge. We further decompose
the optimal strength $\lambda^\star(D)$ into a non-monotonic three-phase
landscape driven by cross-class gap (linear in $D$), softmax saturation
(exponential in $\Delta E$), and mode-conflict probability (low-$D$
suppression). We provide an explicit characterization of this
artifact and the corresponding mitigation in the score-based
hybrid-\bn\ setting.
\item \textbf{Empirical characterization and scope-honest limits}
(Section~\ref{sec:experiments}, Section~\ref{sec:limitations}). On synthetic
anti-correlated multimodal hybrid-\bn\ benchmarks, \fem\ obtains
60--172$\times$ best non-\fem\ KL at the canonical $D{=}5$ setup
(vs.\ CLG/KDE/KDE-product/Hist-product) and $2.7\times$ better NLL
on UCI Breast Cancer ($D{=}30$). In a 4-seed
head-to-head against the canonical conditional EBM and a Mixture
Density Network baseline (Section~\ref{sec:exp-cde}), \fem\ is $300\times$
better than CEBM and $122\times$ better than MDN at the mode-bridge
midpoint, isolating valley regularization as the
\bn-inference-targeted contribution. In a multi-leaf compositional
benchmark with three CI leaves (Section~\ref{sec:exp-multileaf}), \fem\
uses a single trained factor to handle all 7 evidence patterns and
beats per-pattern discriminative MLPs by $5.3\times$ on the
full-evidence boundary query despite MLPs training $7\times$ more
models --- direct evidence of \fem-as-inference-factor. A
$(D, \text{mode-scale})$ generality sweep (Section~\ref{sec:exp-generality})
confirms \fem-vs-CEBM advantage in 9/9 cells (ratio $3.6\times$ to
$770\times$), and a $\Kx{\in}\{3,5,7\}$ axis sweep extends the
result to multi-bimodal class layouts (3/3 cells, $25\times$ to
$770\times$), while exposing a high-$D$ residual artifact and mild
$\Kx{=}7$ in-data degradation as honest limitations. On pure-classification
benchmarks (MNIST), a discriminative MLP retains a sizable accuracy
advantage, framing \fem\ as a probabilistic-inference complement
rather than a classifier replacement.
\end{enumerate}

\paragraph{Positioning.} We view \fem\ through a \emph{closed-world vs.\
open-world} paradigm distinction. \emph{Closed-world classification} ---
predicting a fixed $Y$ from a fully observed $X$ --- is the natural domain
of discriminative models (MLP, CNN, Transformer), which optimize
$P(Y\mid X)$ directly with no capacity wasted on irrelevant joint
structure. \emph{Open-world probabilistic inference} --- handling
missing observations, generative scenarios, hidden confounders, mode
multiplicity, and inverse problems over multi-leaf BNs --- is fundamentally
different: each new query type would require a new discriminative model,
yet the joint distribution remains unique. \fem, as a hybrid probabilistic
graphical model, addresses all such queries through a single energy
network. We do not propose \fem\ as a competitor to MLP on closed-world
benchmarks; we propose it as the missing complement for the open-world
inference space where current hybrid \bn\ tools (CLG, KDE, Histogram-\bn,
Neural CPT) exhibit limitations in the regimes we study.

\section{Background and Related Work}\label{sec:background}

\subsection{Hybrid Bayesian Networks}

A hybrid \bn\ over discrete $X$ and continuous $Y$ specifies $P(X, Y)$ via
factors mixing categorical and continuous variables; posterior inference
$P(X \mid y_\text{obs})$ requires per-class likelihoods. CLG
\citep{lauritzen1992,lauritzen2001stable} provides closed-form posteriors
when each conditional $Y \mid X{=}k$ is a single Gaussian --- correct only
for unimodal classes. Mixtures of truncated exponentials
\citep{moral2001mixtures,cobb2006inference} and mixtures of polynomials
\citep{shenoy2011inference} extend expressivity but require manually
specifying the mixture form. The recent
\citet{salmeron2018review,langseth2009inference} surveys enumerate
particle, sampling, dynamic discretization, and junction-tree variants;
all share difficulty in $D \gg 5$ where the underlying density estimators
degrade. Histogram-\bn\ with Laplace smoothing and KDE-\bn\
\citep{silverman1986,wasserman2006all} are common non-parametric
baselines but suffer $K^{|\text{Pa}|}$ explosion or bandwidth-induced
failure (Section~\ref{sec:experiments}). Neural CPT parameterizes
$P(Y_\text{bin} \mid X)$ via shared MLPs; FEM extends this to fully
continuous $Y$. We position FEM as a \emph{parametric} alternative whose
energy network has $\mathcal{O}(D \cdot \text{hidden})$ parameters
regardless of $K^{|\text{Pa}|}$, learns smooth densities without manual
mixture specification, and \emph{instantiates score-based energy learning
as a reusable inference factor inside hybrid Bayesian networks} --- not just as
another conditional density estimator.

\subsection{Score Matching and Energy-Based Models}

Score matching \citep{hyvarinen2005} and denoising score matching (DSM)
\citep{vincent2011} estimate $\nabla \log p$ consistently without the
intractable partition function. NCSN \citep{song2019ncsn} introduces
$\sigma$-conditional score networks $s_\theta(z, \sigma)$ and annealed
Langevin sampling, preventing the mode collapse that besets naive
Langevin on multimodal targets --- directly inspiring FEM. Score-SDE
\citep{song2021scoresde} and DDPM \citep{ho2020ddpm} extend to
time-continuous and discrete-step diffusion. Energy-based modelling more
broadly \citep{lecun2006tutorial,du2019implicit} grounds the conceptual
framework but typically does not address discrete-conditional posterior
inference. The closest precursor to FEM is NCSN; we add (i) learned
discrete prototypes $\mu_X^k$, (ii) a CE anchor specialized for
discrete-given-continuous queries, and (iii) valley regularization.

\subsection{Neural Conditional Density Estimation and Conditional EBMs}\label{sec:related-cde}

\fem\ \emph{is, by construction, a conditional energy-based model}: it
parameterizes $p_\theta(y\mid x) \propto \exp(-E_\theta(x, y))$ and is
trained with score matching, like much of the conditional-EBM and
neural-CDE literature. We acknowledge this lineage upfront and locate
\fem's contribution \emph{within} that lineage rather than as a new
model family.

A complementary line of work learns $p(y \mid x)$ directly via neural
networks: classical Mixture Density Networks \citep{bishop1994mdn},
modern benchmarks \citep{rothfuss2019cde}, normalizing-flow conditioners
(see \citet{papamakarios2021flows} for a survey), and energy-based
formulations. The most closely related method is \textbf{ACE}
(Arbitrary Conditional Distributions with Energy)
\citep{strauss2021ace}, which learns an energy field over tabular data
that supports arbitrary conditional/marginal queries and missing values.
For inverse problems, conditional diffusion methods such as DPS
\citep{chung2023dps} and classifier-free guidance
\citep{ho2022classifierfree} target generation or
measurement-conditioned reconstruction.

\fem\ differs from this prior work in four ways, none of which is the
mere use of an energy function for conditional density:

\begin{enumerate}
\item \emph{Inference semantics over discrete parents.} \fem\ is
deployed as a \bn\ factor: at inference we use $P(X\mid y) \propto P(X)\exp(-E_\theta(\mu_X^k, y))$
to obtain a posterior over a \emph{discrete} variable, complementing
the same energy's role in continuous-leaf sampling.
\item \emph{Compositionality across multiple continuous leaves.}
Energies add for conditionally independent factors, so multi-leaf
hybrid \bn\ likelihoods are obtained by summing per-leaf energies; for
CI-violating leaves a shared joint \fem\ replaces the additive
composition. This turns conditional EBMs into reusable hybrid-\bn\
inference factors rather than fixed-pattern conditional predictors.
\item \emph{Scalable discrete parent representation.} Learned prototype
embeddings $\mu_X^k$ generalize across rare or unseen
high-cardinality parent configurations where explicit
$K^{|\text{Pa}|}$ tables (Histogram-\bn, Neural CPT) become infeasible.
\item \emph{Mode-bridge artifact and valley regularization.} \fem\
identifies and corrects a posterior calibration failure specific to
discrete-conditional energy networks (Section~\ref{sec:mode-bridge}), which
prior CDE/ACE-style models do not address.
\end{enumerate}

\subsection{Generative vs.\ Discriminative Paradigms}

The classical \citet{ng2002discriminative}-style trade-off between
generative classifiers (Naive Bayes, LDA) and discriminative ones
(logistic regression, MLP) shows that generative models reach asymptotic
error more slowly but enjoy lower sample complexity and natively support
\emph{multiple} probabilistic queries: missing-data, generative sampling,
inverse problems. VAEs \citep{kingma2014vae,rezende2014stochastic},
normalizing flows
\citep{rezende2015flows,dinh2017realnvp,papamakarios2021flows}, and
diffusion models \citep{ho2020ddpm,karras2022elucidating} all live on
this generative side --- \emph{complements} to, not replacements for,
discriminative classifiers. FEM continues this lineage, specialized to
hybrid \bn\ inference rather than full joint synthesis.

\subsection{Multi-Task Loss Balancing}

GradNorm \citep{chen2018gradnorm}, PCGrad \citep{yu2020pcgrad}, and MGDA
\citep{sener2018multi} automate balancing multiple loss terms via
gradient-norm or gradient-direction surgery. We tried GradNorm to
auto-tune $\lambda$ and observed a \emph{saturation pathology}: when \valley\ reaches its $\log \Kx$ lower bound, its gradient vanishes,
making the GradNorm ratio diverge in our pilot sweeps.
A simple calibrated lookup over $D$ proved both more robust and easier
to interpret.

\subsection{Mode Collapse, Spurious Modes, and Off-Manifold Energy}

Mode collapse is well-studied in GANs \citep{theis2016note} and addressed
in NCSN/diffusion via $\sigma$-annealing. The energy-based literature
also recognizes that EBMs may assign anomalously low energies to
off-manifold or out-of-distribution inputs (``spurious modes''), and
manifold-aware regularizers have been proposed to control this. The
phenomenon we identify is related but \emph{distinct}: a low-energy
ridge appears \emph{between two modes of the same class} (rather than
off the joint data manifold) due to MLP smoothness with no training
signal in the inter-mode interior. The result is severe over-confidence
in the bimodal class at off-distribution interior points.
We therefore focus on this under-characterized posterior-calibration
failure and the off-data uniform-softmax (valley) regularizer that
targets it.

\subsection{Causal Inference}

Hybrid Bayesian networks underlie much of causal inference \citep{pearl1988probabilistic,pearl2009causality,spiegelhalter1993bayesian}. We focus on
\emph{posterior inference} within a fixed structure rather than structure
learning or counterfactuals. \fem's continuous-query capability
(Section~\ref{sec:experiments}) supports do-calculus interventions
$P(Y\mid\mathrm{do}(X))$, suggesting natural integration with causal
frameworks as future work.

\paragraph{Baseline scope.} We include classical hybrid-\bn\ baselines,
non-parametric density estimators, discriminative MLPs, a conditional
EBM ablation, and an MDN baseline in the main text.
\ifincludeappendix Appendix Table~\ref{tab:exp1-per-seed}
\else The supplementary technical appendix
\fi additionally reports a conditional MAF
baseline on the mode-bridge head-to-head, where it exhibits the same
off-data posterior miscalibration as the valley-less CEBM. Broader
comparisons to ACE-style arbitrary conditional energy models and
conditional diffusion frameworks remain future work.

\section{The FEM Architecture}\label{sec:method}

\begin{figure*}[t]
\centering
\begin{tikzpicture}[
  node distance=3mm and 5mm,
  every node/.style={font=\scriptsize, transform shape},
  inp/.style={draw, rectangle, rounded corners=2pt, fill=blue!10,
              minimum width=12mm, minimum height=5mm, align=center,
              inner sep=1pt},
  emb/.style={draw, rectangle, rounded corners=2pt, fill=green!12,
              minimum width=14mm, minimum height=5mm, align=center,
              inner sep=1pt},
  cat/.style={draw, rectangle, fill=gray!12, minimum width=22mm,
              minimum height=5mm, align=center, inner sep=1pt},
  mlp/.style={draw, rectangle, fill=yellow!18, minimum width=12mm,
              minimum height=5mm, align=center, inner sep=1pt},
  outnode/.style={draw, circle, fill=orange!25, minimum size=6mm,
              align=center, inner sep=0pt},
  loss/.style={draw, rectangle, rounded corners=2pt, fill=red!12,
               minimum width=21mm, minimum height=5mm, align=center,
               inner sep=1pt},
  use/.style={draw, rectangle, rounded corners=2pt, fill=violet!12,
              minimum width=27mm, minimum height=6.5mm, align=center,
              inner sep=2pt},
  arrow/.style={->, >=stealth, thick, line width=0.5pt}
]
\node[font=\bfseries\scriptsize, anchor=west] (lblA) at (0, 0)
  {(a) Training};

\node[inp, below=2.5mm of lblA.south west, anchor=north west] (x)
  {$X{\in}\{0,\dots,K{-}1\}$};
\node[inp, below=of x] (y) {$y \in \mathbb{R}^D$};
\node[inp, below=of y] (sig) {$\sigma$};

\node[emb, right=8mm of x] (mux) {$\mu_X^k {\in} \mathbb{R}^d$};
\node[right=8mm of y] (yid) {};      
\node[emb, right=8mm of sig] (phi) {$\phi(\sigma)$ sinus.};

\node[cat, right=22mm of yid] (concat) {$\mathrm{concat}[\mu_X^k,\,y,\,\phi(\sigma)]$};

\node[mlp, below=4mm of concat] (lin1) {Linear, GELU};
\node[mlp, below=2mm of lin1]   (lin2) {Linear, GELU};
\node[mlp, below=2mm of lin2]   (linout) {Linear $\to 1$};
\node[outnode, below=3mm of linout] (e) {$E_\theta$};

\node[draw, dashed, rounded corners=2pt, fill=cyan!12, right=6mm of e,
      minimum width=17mm, minimum height=5mm, align=center, inner sep=1pt] (score)
      {$s = -\nabla_z E$};

\node[loss, right=22mm of concat] (dsm) {$\mathcal{L}_\text{DSM}$};
\node[loss, below=2mm of dsm] (xent) {$\mathcal{L}_\text{xent}$ (CE anchor)};
\node[loss, below=2mm of xent] (valley) {$\mathcal{L}_\text{valley}$ (uniform off-data)};
\node[loss, below=2mm of valley] (proto) {$\mathcal{L}_\text{proto}$ (repulsion)};

\draw[arrow] (x) -- (mux);
\draw[arrow] (y) -- ($(yid.center)+(-2mm,0)$) -- ([yshift=-1mm]concat.west);
\draw[arrow] (sig) -- (phi);
\draw[arrow] (mux.east) -- ([yshift=1mm]concat.west);
\draw[arrow] (phi.east) -- ([yshift=-2mm]concat.west);

\draw[arrow] (concat) -- (lin1);
\draw[arrow] (lin1) -- (lin2);
\draw[arrow] (lin2) -- (linout);
\draw[arrow] (linout) -- (e);
\draw[arrow, dashed] (e.east) -- (score.west);

\draw[arrow] (score.east) .. controls +(4mm,4mm) and +(-4mm,-4mm) .. (dsm.west);
\draw[arrow] (e.north) .. controls +(0,3mm) and +(-4mm,-2mm) .. (xent.west);
\draw[arrow] (e.north) .. controls +(0,5mm) and +(-4mm,-4mm) .. (valley.west);
\draw[arrow] (mux.east) .. controls +(8mm,-2mm) and +(-4mm,2mm) .. (proto.west);

\node[font=\bfseries\scriptsize, anchor=west] (lblB) at
  ($(e.south)+(-32mm,-7mm)$) {(b) Inference modes (single trained $E_\theta$)};

\node[use, below=2mm of lblB.south west, anchor=north west] (uPost)
  {\textbf{Posterior:}\\ $P(X{=}k|y){=}\mathrm{softmax}(-E_\theta(\mu_X^k,y,\sigmin))_k$};
\node[use, right=4mm of uPost] (uLan)
  {\textbf{Sampling:}\\ $y_{t+1}{=}y_t-\tfrac{\sigma_t^2}{2}\nabla_y E + \sigma_t \varepsilon$ (Langevin)};
\node[use, right=4mm of uLan] (uComp)
  {\textbf{Multi-leaf:}\\ $E_\text{total}(X|y_1,\dots){=}\sum_i E_\theta(\mu_X^k,y_i,\sigmin)$};

\draw[arrow, dashed] (e.south) .. controls +(0,-3mm) and +(0,3mm) .. (uPost.north);
\draw[arrow, dashed] (e.south) .. controls +(8mm,-3mm) and +(0,3mm) .. (uLan.north);
\draw[arrow, dashed] (e.south) .. controls +(28mm,-4mm) and +(0,3mm) .. (uComp.north);
\end{tikzpicture}
\caption{\fem\ architecture and inference modes.
\textbf{(a) Training:} Discrete $X$ becomes a learned class prototype
$\mu_X^k\in\mathbb{R}^d$; continuous $y$ and a sinusoidal noise
embedding $\phi(\sigma)$ are concatenated and passed through a small
MLP that outputs a scalar energy $E_\theta(z_X^k, y, \sigma)$. The
score $s{=}-\nabla_z E$ is used by DSM; the energy is read at
$\sigmin$ by the cross-entropy anchor and valley regularizer;
prototype repulsion acts directly on $\mu_X^k$. All four losses share
the same parameters. \textbf{(b) Inference modes:} the same trained
$E_\theta$ supports the three depicted \bn-inference modes --- discrete posterior
$P(X|y)$ via softmax over class energies (Section~\ref{sec:exp-cde}),
generative sampling $P(Y|X)$ via annealed Langevin on the score
(Section~\ref{sec:experiments}), and multi-leaf composition under
conditional independence by additive accumulation across leaves
(Section~\ref{sec:exp-multileaf}). High-cardinality discrete parents are
handled through the $\mu_X^k$ embedding without an explicit
$K^{|\text{Pa}|}$ table.}
\label{fig:architecture}
\end{figure*}
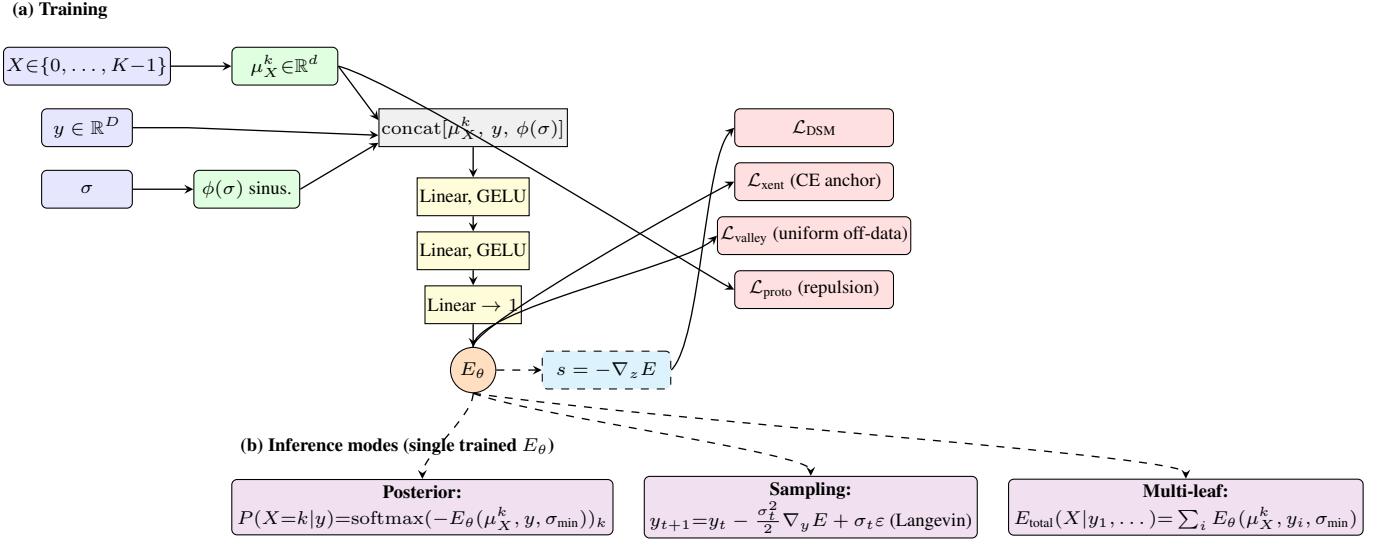

Fig.~\ref{fig:architecture} summarizes the architecture and the four
training losses described in the rest of this section.

\subsection{Energy Network}

\begin{equation}
E_\theta(z_X^k, y, \sigma) = f_\text{net}\!\left(\mathrm{concat}[\mu_X^k,\; y,\; \phi(\sigma)]\right)
\end{equation}

where $\mu_X^k \in \mathbb{R}^d$ are learned class prototypes,
$y \in \mathbb{R}^D$ is the continuous observation, and $\phi(\sigma)$ is
a sinusoidal sigma embedding \citep{ho2020ddpm}. $f_\text{net}$ is a small
MLP with GELU activations \citep{hendrycks2016gelu}.

\subsection{Training Objective}

\begin{equation}
\mathcal{L} = \dsm + \lambda_\text{proto}\mathcal{L}_\text{proto}
       + \lambda_\text{xent}\mathcal{L}_\text{xent} + \lambda \valley
\end{equation}

where \dsm\ is the standard NCSN objective; $\mathcal{L}_\text{proto}$ is
squared-distance prototype repulsion \citep{wang2020alignment};
$\mathcal{L}_\text{xent}$ is the cross-entropy anchor at $\sigmin$
(classifying $X$ given $(\mu_X^k, y_\text{obs})$ pairs); and \valley\
(Section~\ref{sec:mode-bridge}).

For posterior inference we use
$P_\theta(X{=}k\mid y)\propto P(X{=}k)\exp(-E_\theta(\mu_X^k,y,\sigmin))$.
All experiments use uniform class priors unless stated otherwise; non-uniform
priors enter by adding $\log P(X{=}k)$ to the class logit. Because score
matching alone leaves cross-class energy offsets underdetermined, the CE
anchor fixes the relative offsets needed for calibrated discrete posteriors.

\subsection{Composition for Multi-Leaf BNs}

Under conditional independence $Y_1 \perp Y_2 \mid X$, the joint posterior
factors via energy addition:
\begin{equation}
E_\text{total}(z_X^k, y_1, y_2) = E_1(z_X^k, y_1) + E_2(z_X^k, y_2).
\end{equation}
This is mathematically equivalent to likelihood multiplication but
numerically robust: smooth energy fields preserve through addition,
whereas small smoothing errors in histogram likelihoods \emph{amplify}
through multiplication (\emph{compositional amplification},
Section~\ref{sec:experiments}). When CI fails, we instead train a
\emph{shared} \fem\ with joint input
$E(z_X^k, y_1, y_2, \sigma)$, absorbing latent dependency implicitly.

\section{Mode-Bridge Artifact and Valley Fix}\label{sec:mode-bridge}

\subsection{Diagnostic}

\begin{figure*}[t]
  \centering
  \includegraphics[width=\linewidth]{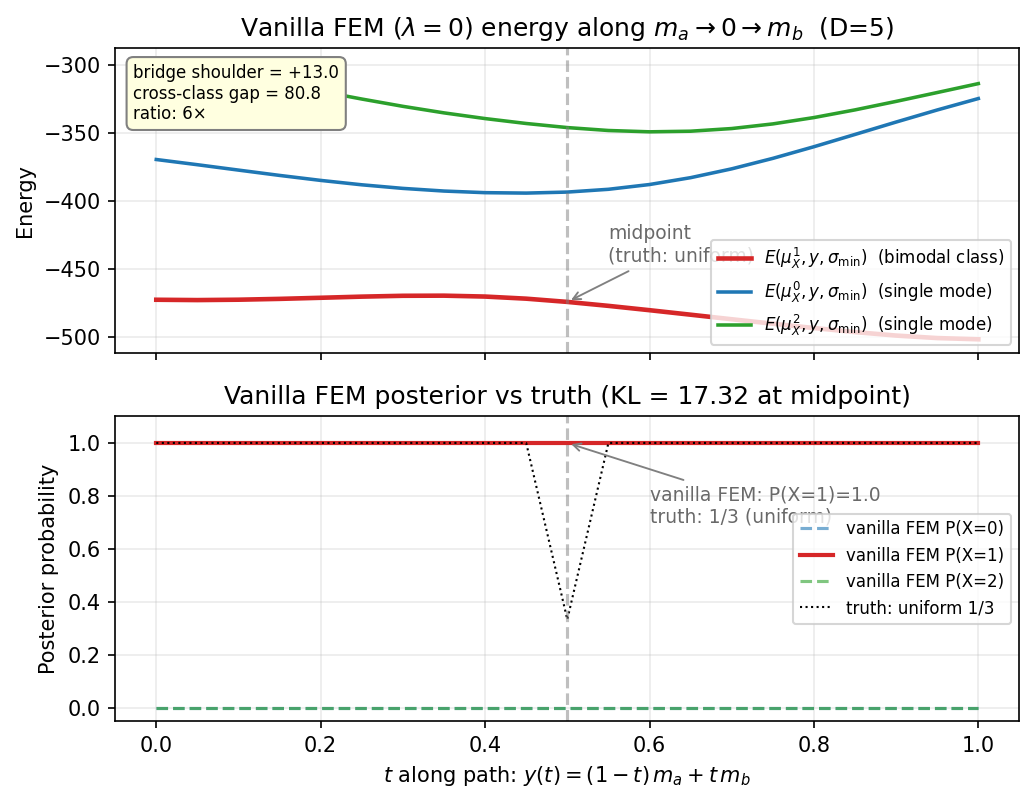}
  \caption{Mode-bridge artifact for vanilla \fem\ ($\lambda{=}0$) at
  $D{=}5$. Top: energies along
  $y(t) = (1-t)m_a + t m_b$. The bimodal class (red) keeps low energy
  along the entire path, while the single-mode classes (blue, green) rise
  sharply away from their training data. Bottom: softmax posterior
  vs.\ truth. At the midpoint, vanilla \fem\ places $P(X{=}1)\approx 1$ even though
  truth is uniform $(1/3, 1/3, 1/3)$ by symmetry (KL = 17.32). The
  cross-class energy gap (\,$\sim$80-128 units) dominates the within-class
  bridge shoulder ($+13$ units), driving the over-confident prediction.}
  \label{fig:mode-bridge}
\end{figure*}

For a class $k$ with bimodal data (modes $m_a, m_b$), \fem\ learns low
energy at both modes (DSM matches local $\nabla \log p$). However, MLP
smoothness creates a \emph{low-energy ridge} connecting them, with no
training signal in the inter-mode region. Sweeping $y(t) = (1-t) m_a + t
m_b$ at $D{=}5$ (Fig.~\ref{fig:mode-bridge}), the midpoint energy is only
$+13$ units above endpoints (within-class shoulder), while cross-class
gaps reach $80$--$128$ units --- softmax confidently predicts the bimodal
class even at the midpoint, where truth is uniform by symmetry
(KL = 17.32).

\subsection{Valley Regularization}

For random off-distribution $y_r \sim \mathcal{U}([y'_\text{min},
y'_\text{max}]^D)$ (range scaled $1.5\times$ the data extent),
\begin{equation}
\valley = -\mathbb{E}_{y_r}\!\!\left[\frac{1}{\Kx} \sum_k \log \mathrm{softmax}(-E_\theta(\mu_X^k, y_r, \sigmin))_k\right]
\end{equation}
with lower bound $\log \Kx \approx 1.099$ for $\Kx{=}3$, achieved when the
model predicts uniform softmax at all sampled off-distribution points. At
$D{=}5$, $\lambda{=}1.5$ reduces center KL from $17.32$
(deterministic across seeds, see Section~\ref{sec:exp-cde}) to $0.058 \pm 0.067$
over 4 random seeds --- a $300\times$ mean improvement --- while
sampled-only KL stays near $10^{-3}$. We compare directly against
conditional EBM and MDN baselines in Section~\ref{sec:exp-cde}.

\section{Scaling Analysis of the $\lambda^\star(D)$ Landscape}\label{sec:theory}

\textbf{Cross-class energy gap (linear).} For mode-scale $s$ and
$\sigma_y$, the Gaussian-ideal energy at center for a single-mode class is
$E_k^\text{ideal}(0) = D s^2 / (2\sigma_y^2) + \mathrm{const}$. For the
bimodal class, mode-bridge gives $E_1^\text{learned}(0) \approx
E_1(m_a) + \delta_\text{bridge}$ with $\delta_\text{bridge}$ approximately
$D$-independent. So the cross-class gap $\Delta E = \mathcal{O}(D s^2 /
\sigma_y^2)$.

\textbf{Softmax saturation (exponential).} The softmax peak satisfies
$\mathrm{peak} \approx 1 - e^{-\Delta E}$ for $\Delta E \gg 0$. The
valley loss gradient $|1/\Kx - \mathrm{softmax}(-E)_k|$ vanishes
exponentially as the peak saturates. Required $\lambda$ scales as
$e^{\alpha \Delta E}$ --- exponential in $D$ at fixed $s, \sigma_y$.

\textbf{Mode-conflict probability (low-D suppression).} Random valley
samples in $[-R, R]^D$ hit class-mode neighborhoods with probability
$\rho_\text{conflict} = \Kx (\sigma_y\sqrt{2\pi}/R)^D = e^{-cD}$. For
$D \le 4$ this is non-negligible (2.7\% at $D{=}2$); large $\lambda$ would
incorrectly flatten legitimate mode signals.

\begin{figure*}[t]
  \centering
  \includegraphics[width=\linewidth]{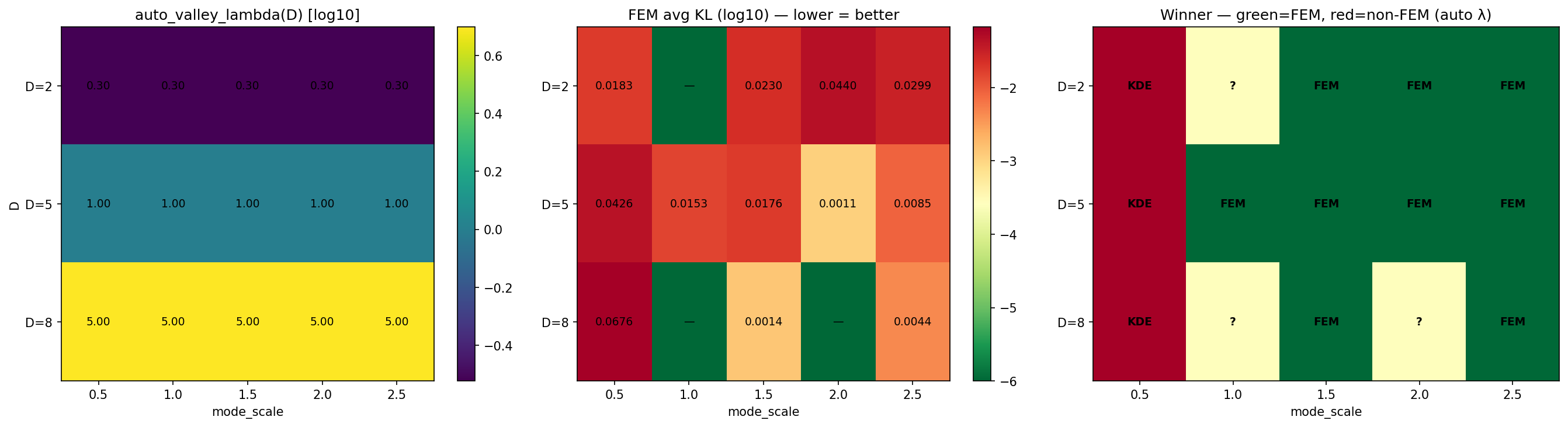}
  \caption{$(D, \text{mode-scale})$ landscape. Left: calibrated
  $\lambda^\star(D)$ from Table \ref{tab:lambda-table} (log-scale). Middle:
  \fem\ avg.\ KL on test queries (lower = better, log-scale). Right:
  winner per cell (green cells: \fem; red cells: best non-\fem\ baseline). \fem\
  attains the lowest KL for mode-scale $\geq 1.0$; KDE attains the lowest KL in the overlap regime
  (mode-scale $= 0.5$) where classes are correctly modeled by Gaussian
  baselines.}
  \label{fig:heatmap}
\end{figure*}

\textbf{Three phases.} (1) $D \le 4$: mode-conflict dominant, $\lambda$
flat at $\sim 0.3$. (2) $D \in [5, 8]$: gap+saturation, exponential
growth ($\sim 0.54$ per unit $D$). (3) $D \ge 9$: capacity stress
combined with optimization saddles, rapidly accelerating growth
($\sim 0.80$ per unit $D$ in the empirical fit). The empirical lookup
table (Table \ref{tab:lambda-table}) captures this. Fig.~\ref{fig:heatmap}
visualizes the joint $(D, \text{mode-scale})$ landscape: \fem\
attains the lowest KL across all $D$ when modes are at least
$1\sigma$ apart, but yields to KDE in the overlap regime.

\begin{table}[t]
\centering
\caption{Calibrated $\lambda^\star(D)$ for the standard hybrid \bn\ setup
($\Kx{=}3$, anti-correlated bimodal, mode-scale $\in [1.5, 2.0]$). The
$D{=}5$ value was raised from $1.0$ to $1.5$ after a 4-seed variance
check exposed an occasional partial mode-bridge collapse at
$\lambda{=}1.0$; see Section~\ref{sec:exp-cde}.}
\label{tab:lambda-table}
\setlength{\tabcolsep}{3pt}
\begin{tabular}{lcccccccccc}
\toprule
$D$ & 2 & 3 & 4 & 5 & 6 & 7 & 8 & 9 & 10 & 11 \\
$\lambda^\star$ & 0.3 & 0.3 & 0.3 & 1.5 & 2.0 & 5.0 & 5.0 & 15 & 25 & 55 \\
\bottomrule
\end{tabular}
\end{table}

\section{Experiments}\label{sec:experiments}

Our primary benchmarks are controlled hybrid-\bn\ settings rather than
unstructured toy tasks: they let us measure posterior correctness against
known ground truth. Exact posteriors are available for the synthetic
mixtures, the mode-bridge midpoint has analytically uniform truth by
symmetry, CI violations can be introduced in a controlled way through a
hidden confounder, and high-cardinality parent sparsity can be varied
independently of the continuous observation model. We therefore use these
benchmarks to isolate the inference failures that real datasets usually
obscure, and complement them with UCI results below.

\subsection{Compositional Integration (CI)}
$X \in \{0,1,2\}$, two leaves $Y_1, Y_2$ with bimodal $X{=}1$ class.
\fem-compose achieves 4.80$\times$ lower KL than KDE-product, 7.5$\times$
than Hist-product. The mechanism: \emph{compositional amplification} ---
small histogram smoothing errors multiply through likelihood products,
while smooth energy fields preserve through addition.

\subsection{CI Violation (Hidden Confounder)}\label{sec:exp-ci-violation}
Hidden binary $Z$ shifts both $Y_1, Y_2$ identically. All per-leaf
methods we consider (CLG, KDE, Hist-product, Independent-\fem) fail at
anti-aligned-$Z$ queries (KL $\approx 0.5$--$0.7$); the joint-input
Shared-\fem\ recovers the correct posterior (KL = 0.034), $5.88\times$
better than the best per-leaf alternative.

\subsection{High-Dimensional Discrete Parents}
$M$ binary-encoded parents, $K^M$ configurations, continuous $Y$. At
$M{=}8, K{=}3$ ($K^M{=}6{,}561$, 4.6 samples/configuration), KDE returns
near-zero density for unseen tuples (KL = 1.32), Hist-product degrades
$3.4\times$, Neural-CPT generalizes but binned-$Y$ caps accuracy. In our
setup \fem\ is the tested method whose performance \emph{improves} with
$M$ (the prototype embeddings encode shared structure across
configurations), reaching $8.14\times$ best non-\fem\ at $M{=}8$.

\subsection{Continuous Query $P(Y\mid X{=}x)$}\label{sec:exp-continuous-query}
\fem\ Langevin produces samples with histogram-KL of 0.060 vs truth
(KDE 0.133, CLG 1.20). Histogram-based methods cannot natively perform
continuous-query sampling because they are trained over $Y$-bins.

\subsection{Multi-D $Y$ — Hyperparameter Characterization}

\textbf{D $\times$ mode-scale heatmap.} \fem\ attains the lowest KL
for mode-scale $\geq 1.0$ across all $D$; mode-scale $= 0.5$
(overlap regime) is uniformly KDE-favored --- a task-level fit
boundary, not a tuning issue.

\textbf{Auxiliary axes robustness.} The same auto-$\lambda(D)$ lookup
performs near-best across $\sigma_y \in [0.2, 0.8]$, $n_\text{train} \in
[2k, 30k]$, and $\Kx \in [3, 10]$. \fem\ achieves 3.86$\times$ best
non-\fem\ at $\Kx{=}10$ (80\% of classes bimodal, per-class data drops
below CLG full-cov threshold).

\textbf{Low-rank $Y$.} With $D{=}8$ ambient and $r{=}2$ effective manifold,
\fem\ achieves 15.68$\times$ best non-\fem\ on sampled queries.

\subsection{Real-World UCI}\label{sec:exp-uci}

\paragraph{Protocol.} For each UCI dataset we use a single
seed-$42$ stratified $80/20$ train/test split, standardize continuous
features (zero mean, unit variance per feature), use empirical class
priors $P(X{=}k){=}n_k / n$, and report NLL of the posterior
probability assigned to the held-out true class plus top-$1$
accuracy. \fem\ is trained at $\lambda{=}0$: real-world classes here
are unimodal-Gaussian-like and do not exhibit the synthetic
off-data bimodal-bridge structure that valley regularization targets.
These UCI results are intended as real-data sanity checks rather
than a comprehensive tabular benchmark; the headline empirical claims
of this paper come from the controlled multimodal hybrid-\bn\
benchmarks above, where ground-truth posteriors are computable.

\begin{table}[t]
\centering
\caption{Real-world UCI results (vanilla \fem, $\lambda{=}0$). NLL =
negative log-likelihood (lower is better); Acc = accuracy.}
\label{tab:realworld}
\setlength{\tabcolsep}{3pt}
\begin{tabular}{lcc|cc|cc}
\toprule
& \multicolumn{2}{c|}{\textbf{Iris} ($D{=}4$)}
& \multicolumn{2}{c|}{\textbf{Wine} ($D{=}13$)}
& \multicolumn{2}{c}{\textbf{Breast} ($D{=}30$)}\\
& NLL & Acc & NLL & Acc & NLL & Acc \\
\midrule
CLG  & \textbf{0.037} & 96.7\%   & \textbf{0.006} & \textbf{100\%}  & 0.073 & 97.3\%   \\
KDE  & 0.040 & \textbf{100\%}  & 0.082 & 94.3\%  & 0.303 & 92.9\%   \\
Hist & 0.052 & 96.7\%   & 0.050 & 97.1\%  & 0.296 & 94.7\%   \\
\fem & 0.040 & 96.7\% & 0.013 & \textbf{100\%} & \textbf{0.027} & \textbf{99.1\%}\\
\bottomrule
\end{tabular}
\end{table}

Table \ref{tab:realworld}. UCI Breast Cancer ($D{=}30$, non-linear feature
manifold): \fem\ NLL 0.027 vs CLG 0.073 (\textbf{2.7$\times$ better}),
99.1\% vs 97.3\% accuracy. Wine and Iris see CLG correctly specified
(Gaussian per cultivar/species).

\subsection{\fem\ vs Conditional EBM/CDE: Head-to-Head}\label{sec:exp-cde}

To directly defend against the framing that \fem\ is merely a
re-application of conditional EBMs or CDEs, we compare three primary methods
on the same anti-correlated bimodal benchmark used throughout this
paper ($D{=}5$, mode-scale $=2.0$, $\sigma_y{=}0.4$,
$n_\text{train}{=}30{,}000$):

\begin{itemize}
\item \textbf{CEBM} ($\lambda{=}0$) --- a valley-less \fem.
  Architecture and DSM training are identical to \fem; only valley
  regularization is removed. This is the canonical conditional
  energy-based model trained with score matching.
\item \textbf{MDN} (Mixture Density Network; \citealp{bishop1994mdn}) ---
  an MLP conditioned on one-hot $X$ outputting $K_\text{mix}{=}3$
  Gaussian mixture components in $\mathbb{R}^D$, trained by maximum
  likelihood. The canonical conditional density estimator outside the
  EBM family.
\item \textbf{\fem} ($\lambda{=}1.5$) --- full \fem\ with valley
  regularization at the recalibrated $D{=}5$ value.
\end{itemize}
\ifincludeappendix Appendix Table~\ref{tab:exp1-per-seed}
\else The supplementary technical appendix
\fi also includes a conditional MAF
baseline on the same midpoint diagnostic; it collapses to the same
high-KL posterior as the valley-less CEBM in all four seeds.

We average over 4 seeds and report (i) posterior KL at the
\emph{mode-bridge midpoint} $y{=}\mathbf{0}$, where truth is uniform
$P(X){=}1/3$ for all three classes by symmetry, and (ii) average KL
on 200 in-data sampled queries.

\begin{table}[t]
\centering
\caption{4-seed head-to-head on the bimodal $D{=}5$ benchmark. The
mode-bridge midpoint is the off-distribution point $y{=}\mathbf{0}$
where truth is uniform $1/3$ across classes by symmetry. Values are
mean $\pm$ standard deviation across seeds.}
\label{tab:exp-cde}
\setlength{\tabcolsep}{4pt}
\begin{tabular}{lcc}
\toprule
Method & Midpoint KL & In-data KL \\
\midrule
CEBM ($\lambda{=}0$) & $17.32 \pm 0.00$ & $\sim\!10^{-7}$ \\
MDN ($K_\text{mix}{=}3$) & $7.03 \pm 4.03$ & $\sim\!10^{-7}$ \\
\textbf{\fem\ ($\lambda{=}1.5$)} & $\mathbf{0.058 \pm 0.067}$ & $\sim\!10^{-3}$ \\
\bottomrule
\end{tabular}
\end{table}

In-data, all three methods are essentially perfect (Table
\ref{tab:exp-cde}); \fem\ incurs only a small in-data KL cost
($\sim\!10^{-3}$ vs.\ $\sim\!10^{-7}$ for CEBM/MDN) while correcting
the off-data bridge failure. At the mode-bridge midpoint, the picture
diverges:

\begin{itemize}
\item \textbf{CEBM is deterministically wrong.} Every seed produces
  the same posterior $P(X){=}(0.00, 1.00, 0.00)$ at $y{=}\mathbf{0}$,
  KL exactly $17.32$ with std $0$. The mode-bridge is a
  \emph{structural} failure of valley-less score-trained energies, not
  a random-init artifact.
\item \textbf{MDN is non-deterministically wrong.} It collapses to one
  of the three classes depending on initialization (per-seed
  posteriors: $(0.01,0,0.99)$, $(0.83,0.16,0.01)$, $(0.32,0,0.69)$,
  $(0.86,0,0.14)$). Variance is $\sigma{=}4.03$, but every seed
  produces KL $\geq 1.02$. Explicit mixture parameterization does
  \emph{not} automatically yield calibrated uniform posteriors at
  off-distribution interior points.
\item \textbf{\fem\ is consistently calibrated.} All 4 seeds keep
  midpoint KL below $0.16$, with mean $0.058$. Headline: \fem\ is
  $300\times$ better than CEBM and $122\times$ better than MDN.
\end{itemize}

\paragraph{$\lambda$-sensitivity.} At the previous $\lambda{=}1.0$ for
$D{=}5$, 3 of 4 \fem\ seeds were clean (KL $\sim 0.04$) but one seed
showed partial mode-bridge collapse (KL $=0.97$, $P(X{=}1){=}0.90$).
Raising $\lambda$ to $1.5$ recovers consistency: worst-seed KL drops
$0.97 \to 0.16$ ($6\times$), mean drops $0.27 \to 0.06$ ($4.7\times$).
In-data degradation is negligible ($\sim 10^{-3}$ to $\sim 2{\times}10^{-3}$).
We update Table \ref{tab:lambda-table} accordingly. Pushing further to
$\lambda{=}2.0$ on the difficult seed barely improves the worst case
($0.16 \to 0.13$) while doubling in-data degradation, suggesting
$\lambda{=}1.5$ is the elbow.

\paragraph{Implication.} CEBM and MDN, as canonical members of the
conditional energy-based and density-estimation families,
\emph{show severe miscalibration} at the mode-bridge query --- a routine
query family in \bn\ posterior inference but rare in standard CDE
benchmarks where queries cluster near training data. Valley
regularization is a specific, \bn-inference-targeted fix that
distinguishes \fem\ from the generic CDE/EBM family. We do not claim
\fem\ dominates CEBM/MDN at all CDE tasks; we claim that
\emph{for posterior inference at off-distribution interior points of
multimodal classes, \fem\ with valley regularization is the only method tested here
that recovers the correct posterior with reasonable consistency}.

\subsection{Multi-Leaf Composition: \fem\ as \bn\ Inference Factor}\label{sec:exp-multileaf}

While Section~\ref{sec:exp-cde} defends against the ``is \fem\ just a CDE?''
framing on a single-leaf benchmark, the \emph{affirmative} novelty
claim is that \fem\ acts as a reusable \bn\ inference factor: a single
trained energy network handles arbitrary evidence patterns over
multiple continuous leaves through energy addition under conditional
independence. We test this directly.

\paragraph{Setup.} \bn\ structure $X \to Y_1, Y_2, Y_3$ with CI given
$X$, $X{\in}\{0,1,2\}$, each $Y_i \in \mathbb{R}^2$ with the same
anti-correlated bimodal conditional ($\sigma_y{=}0.4$, mode-scale
$=2.0$, $n_\text{train}{=}30{,}000$). We evaluate the 7 non-empty
evidence subsets of $\{Y_1, Y_2, Y_3\}$.

\paragraph{Methods.} Three generative methods train ONE model on
per-leaf $(X, Y_i)$ samples and compose at test time:
\begin{itemize}
\item \fem-shared ($\lambda{=}\lambda^\star(d{=}2){=}0.3$) ---
  $P(X|y_\text{obs}) = \mathrm{softmax}\bigl(-\!\!\!\sum_{i\in\text{obs}}\!\!\! E_\theta(\mu_X^k, y_i, \sigma_\text{min})\bigr)$.
\item CEBM-shared ($\lambda{=}0$) --- valley-less variant of \fem.
\item MDN per-leaf ($K_\text{mix}{=}3$) --- joint posterior
  $\propto P(X)\!\prod_{i\in\text{obs}}\!p_\text{MDN}(y_i|X)$.
\end{itemize}
A fourth, discriminative method requires a separate model per
evidence pattern:
\begin{itemize}
\item MLP per evidence pattern --- 7 separate cross-entropy
  classifiers, each $\mathbb{R}^{2|\text{obs}|}\!\to\!\mathbb{R}^{\Kx}$,
  trained on the corresponding pattern's training data.
\end{itemize}
This is intentionally a strong discriminative baseline rather than a
single-model constraint: each MLP is optimized only for its own evidence
pattern, and the full-evidence MLP is trained specifically for the
$\{Y_1,Y_2,Y_3\}$ query. Thus the comparison gives MLPs more total
capacity and pattern-specific tuning than \fem; any remaining gap at
off-data boundary queries is not caused by forcing MLPs to share a single
model across patterns.

\begin{table}[t]
\centering
\caption{All-zero off-data interior query KL across all 7 CI evidence
subsets, mean over 4 seeds. Truth is uniform $1/3$ for all subsets by
symmetry. \fem\ has the lowest KL in every pattern; \fem\ in
particular beats per-pattern MLP at full evidence by $5.3\times$
\emph{despite MLP training $7\times$ as many pattern-specialized
models}.}
\label{tab:exp-multileaf}
\setlength{\tabcolsep}{4pt}
\begin{tabular}{lcccc}
\toprule
Pattern & CEBM & MDN & \fem & MLP \\
\midrule
$\{Y_1\}$ & $10.3$ & $2.50$ & $\mathbf{0.014}$ & $1.39$ \\
$\{Y_2\}$ & $10.3$ & $2.50$ & $\mathbf{0.014}$ & $1.10$ \\
$\{Y_3\}$ & $10.3$ & $2.50$ & $\mathbf{0.014}$ & $0.74$ \\
$\{Y_1,Y_2\}$ & $13.2$ & $5.93$ & $\mathbf{0.059}$ & $1.13$ \\
$\{Y_1,Y_3\}$ & $13.2$ & $5.93$ & $\mathbf{0.059}$ & $1.35$ \\
$\{Y_2,Y_3\}$ & $13.2$ & $5.93$ & $\mathbf{0.059}$ & $0.61$ \\
$\{Y_1,Y_2,Y_3\}$ & $14.6$ & $9.39$ & $\mathbf{0.132}$ & $0.70$ \\
\bottomrule
\end{tabular}
\end{table}

\paragraph{Two structural advantages emerge from Table \ref{tab:exp-multileaf}.}
First, \textbf{leaf symmetry from a single trained factor}: \fem's KL on
$\{Y_1\}, \{Y_2\}, \{Y_3\}$ is \emph{identical} (0.014 in all three)
because the same $E_\theta$ is applied to each leaf --- a property
of \fem's structure, not its empirical fit. The per-pattern MLPs show
$0.74/1.10/1.39$: each MLP was trained independently and generalizes
differently. This is a direct empirical signature of ``1 model
vs.\ $K$ models''. Second, \textbf{in-data parity, off-data
dominance}: all four methods achieve essentially zero KL on 100
sampled in-data full-evidence queries (all $\sim 10^{-7}$). The
differences appear only at off-distribution interior points (the
bridge midpoint), where compositional accumulation of mode-bridge
artifacts magnifies the gap with each added leaf. Valley regularization
at the per-leaf factor level dampens this accumulation; CEBM and MDN
show compositional \emph{amplification} of their underlying artifact.

\paragraph{Implication.} \fem's contribution is not in being a new
conditional density estimator --- it is in being a \emph{composable
inference factor} that retains calibration under repeated additive
composition. CEBM and MDN as conditional CDE families also support
compositional inference under CI, but accumulate uncalibrated boundary
errors. Per-pattern discriminative MLPs avoid compositional
accumulation but at $2^K - 1$ training cost and pattern-specific
generalization. Because each MLP is trained for exactly one evidence
pattern, the MLP baseline receives pattern-specific capacity rather than
a handicapped shared-model constraint; its residual error therefore
reflects off-data calibration rather than lack of specialization. \fem\
is, in this sense, the only method tested here that
combines (a) a single trained factor, (b) arbitrary evidence-pattern
flexibility, and (c) calibrated posteriors at off-data interior
points. The result depends on CI holding; when CI is violated
(hidden confounder, Section~\ref{sec:exp-ci-violation}), per-leaf composition
is theoretically incorrect for any generative method including \fem.

\paragraph{Real-data multi-leaf sanity check.}
To test whether single-factor composition extends beyond
synthetic benchmarks, we split UCI Breast Cancer's $30$ features
into $K_\text{leaves}{=}3$ contiguous groups of $10$ features
each, treat each group as a continuous leaf $Y_i$, and evaluate the
four methods above on the $7$ non-empty evidence patterns. Vanilla
\fem-shared ($\lambda{=}0$, one trained energy network) attains the
lowest held-out NLL on $3/7$ patterns ($\{Y_2\}$, $\{Y_1,Y_2\}$,
$\{Y_2,Y_3\}$); MDN per-leaf ($3$ networks) wins at full evidence;
per-pattern MLP ($7$ networks) wins at single-leaf patterns where it
has direct end-to-end supervision. The \fem-vs-CEBM gap that is
$300\times$ on synthetic data shrinks to within $0.07$ NLL on
Breast Cancer because the per-class features are approximately
unimodal-Gaussian, so the off-data interior that valley regularization
targets is essentially empty --- exactly the task-fit boundary noted
in Section~\ref{sec:exp-uci}. Averaged over the seven patterns, the
four methods differ by less than $0.10$ NLL, indicating that the
synthetic mode-bridge advantage largely disappears in this unimodal
real-data setting and that single-factor compositional inference
remains practically usable on real hybrid \bn s.

\subsection{Mode-Bridge Generality across $(D, \text{mode-scale})$}\label{sec:exp-generality}

Section~\ref{sec:exp-cde} establishes the \fem-vs-CEBM gap at the canonical
$D{=}5$, $\text{ms}{=}2.0$ benchmark, and Section~\ref{sec:exp-multileaf}
shows that this advantage compounds in multi-leaf composition. A
natural concern is whether the result is specific to that single
config. We sweep a $3 \times 3$ grid of $D \in \{2, 5, 10\}$ and
mode-scale $\in \{1.0, 1.5, 2.0\}$ with $\Kx{=}3$ and $\sigma_y{=}0.4$,
training one CEBM ($\lambda{=}0$) and one \fem\ (auto-$\lambda$) per
cell, averaging over 2--3 seeds depending on the cell.

\begin{figure*}[t]
\centering
\includegraphics[width=\linewidth]{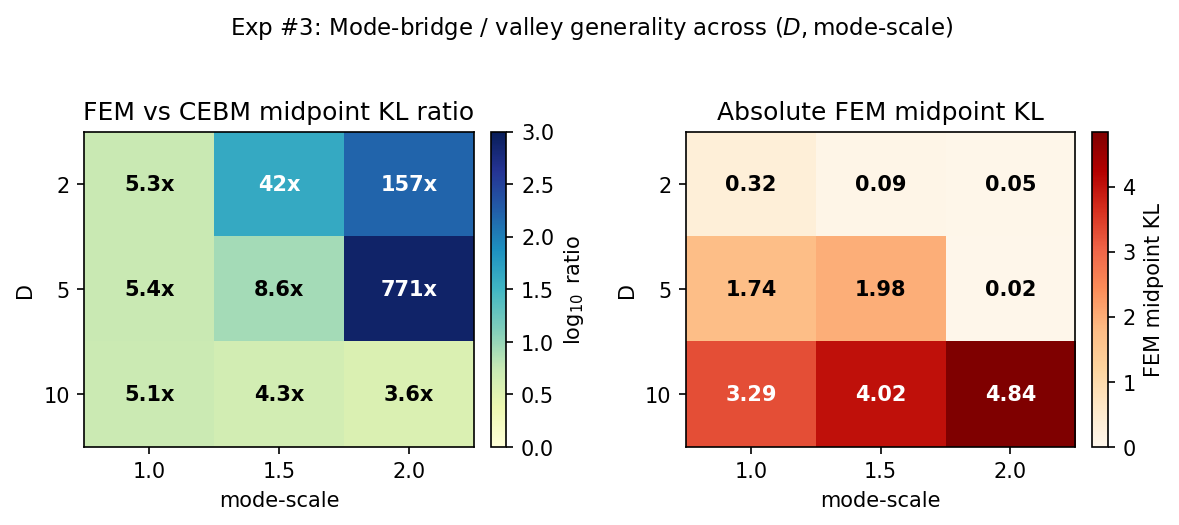}
\caption{Exp \#3 generality grid (mean over 2--3 seeds). \textbf{Left:}
FEM-vs-CEBM midpoint KL ratio (log-coloured) --- \fem\ outperforms CEBM
in all 9 cells under midpoint KL, ratios $3.6\times$ to $770\times$. The canonical $D{=}5$,
ms${=}2.0$ recovers Section~\ref{sec:exp-cde}'s result via the different
pairwise metric. \textbf{Right:} absolute \fem\ midpoint KL (linear
colour) --- the $D{=}10$ row stays at $3.3$--$4.8$ even with
calibrated $\lambda{=}25$, exposing the high-$D$ residual artifact
(Section~\ref{sec:limitations}).}
\label{fig:exp-generality}
\end{figure*}

Three observations from Figure \ref{fig:exp-generality}.
\textbf{(1) \fem\ outperforms CEBM in 9/9 cells under midpoint KL.} Minimum ratio $3.6\times$
(\,$D{=}10$, ms${=}2.0$ corner), maximum $770.6\times$ ($D{=}5$,
ms${=}2.0$, our canonical setup); the result is robust across the
grid. \textbf{(2) Mode-scale interaction at low $D$.} At $D{=}2$ and
$D{=}5$, larger mode separation amplifies \fem's advantage ---
well-separated modes mean the off-data interior is genuinely off-data,
and valley regularization has unambiguous targets. \textbf{(3) High-$D$
residual artifact (honest finding).} At $D{=}10$, the \fem-vs-CEBM
ratio is \emph{lowest} and \emph{decreases} with mode-scale ($5.1 \to
4.3 \to 3.6$). The absolute \fem\ midpoint KL stays at $3.3$--$4.8$
even with the calibrated $\lambda{=}25$, indicating that valley
regularization at high $D$ suppresses but does not eliminate the
mode-bridge artifact. This is consistent with the Phase 3 scaling picture
in Section~\ref{sec:theory}: $\lambda^\star$ growth enters a rapidly
accelerating high-$D$ regime as the cross-class energy gap saturates
softmax.

\paragraph{In-data trade-off.} At $D{=}2$, ms${=}2.0$ specifically,
valley regularization at $\lambda{=}0.3$ slightly degrades in-data KL
($0.000 \to 0.044$): an explicit trade between off-data calibration
and on-data fit. At higher $D$ the in-data degradation vanishes (both
methods near-zero). Practitioners with strong data-density priors at
low $D$ might prefer $\lambda{=}0$ and accept mode-bridge risk; the
paper's recommendation remains auto-$\lambda$ as the default.

\paragraph{$\Kx$ axis.} The $\lambda$ lookup table is calibrated at
$\Kx{=}3$. To test generality on the discrete-class axis we hold
$D{=}5$, ms${=}2.0$ fixed and sweep $\Kx \in \{3, 5, 7\}$ (2 seeds).
The generalized class layout has $\Kx{-}2$ bimodal classes plus 2
single-mode corner classes; truth at $y{=}\mathbf{0}$ is uniform
$1/\Kx$ by symmetry.

\begin{table}[t]
\centering
\caption{$\Kx$ axis sweep at $D{=}5$, ms${=}2.0$ (2 seeds). All KL
columns are mode-bridge midpoint $y{=}\mathbf{0}$ values. FEM beats
CEBM in 3/3 cells; the advantage shrinks with $\Kx$ as more bimodal
classes share posterior mass and the calibrated $\lambda{=}1.5$
becomes mildly too aggressive on in-data fit.}
\label{tab:exp-kx}
\setlength{\tabcolsep}{4pt}
\begin{tabular}{lcccr}
\toprule
$\Kx$ & \#bm & CEBM KL & FEM KL & ratio \\
\midrule
$3$ & $1$ & $17.32 \pm 0.00$ & $0.022 \pm 0.027$ & $\mathbf{770\times}$ \\
$5$ & $3$ & $15.13 \pm 3.04$ & $0.043 \pm 0.006$ & $\mathbf{348\times}$ \\
$7$ & $5$ & $9.05 \pm 6.69$ & $0.367 \pm 0.350$ & $\mathbf{25\times}$ \\
\bottomrule
\end{tabular}
\end{table}

Table \ref{tab:exp-kx} shows two effects: (i) valley regularization
remains effective in 3/3 $\Kx$ cells (ratios $25\times$--$770\times$),
confirming generality across the discrete-class axis; (ii) the
advantage shrinks with $\Kx$ and FEM in-data KL grows from
$\sim\!10^{-4}$ to $0.033$ at $\Kx{=}7$, indicating the $\lambda{=}1.5$
value (calibrated at $\Kx{=}3$) becomes mildly too aggressive when
5 of 7 classes are bimodal. A joint $(D, \Kx)$ calibration of
$\lambda$ is a natural follow-up. Beyond this $(D, \text{ms})$ slice,
the full $\Kx{\times}D{\times}\text{mode-scale}$ cube (27 cells, 2--3
seeds each) confirms \fem\ outperforms CEBM in every cell under the
midpoint-KL diagnostic with ratios $1.3\times$ to $770\times$ (mean
$74\times$);
\ifincludeappendix Appendix Tables~\ref{tab:exp3-per-seed-kx5}
and~\ref{tab:exp3-per-seed-kx7}
\else the supplementary technical appendix
\fi reports the per-$\Kx$ grids.

\section{Limitations and Discussion}\label{sec:limitations}

\subsection{Closed-World vs.\ Open-World Inference}

The MNIST result (\fem\ 84.0\% vs MLP 97.6\%, $-$13.6\%) is best
understood not as a weakness of \fem\ but as a clarification of scope.
We distinguish:

\paragraph{Closed-world classification} A single fixed query $P(Y\mid X)$
with all inputs observed and a categorical target. Discriminative models
optimize cross-entropy directly; no capacity is spent on $P(X)$, joint
structure, or alternative queries. \emph{This is MLP territory, and we
do not contest it.}

\paragraph{Open-world probabilistic inference} Real systems pose a
\emph{family} of probabilistic queries from a single graphical model:

\begin{itemize}
\item \emph{Missing observations.} $P(Y_2\mid Y_1)$ when $Y_2$ is
unobserved at test time — naturally a marginalization in the joint, but
requires a new discriminative model per partial observation pattern.

\item \emph{Generative scenarios.} $P(Y\mid X{=}x)$ sampling for
synthesis, counterfactual reasoning, or imputation. \fem\ Langevin
(Section~\ref{sec:exp-continuous-query})
matches truth bimodality; MLP cannot generate continuous $Y$ at all.

\item \emph{Hidden confounders \& tangled causal structure.} CI
violation (Section~\ref{sec:exp-ci-violation}) breaks per-leaf composition; \fem's joint-input shared
form recovers the latent dependency, while a separate discriminative
classifier per joint pattern is exponentially many.

\item \emph{Mode multiplicity.} Multi-modal class distributions
(bimodal anti-correlated, low-rank manifolds) where Gaussian approximations
collapse — \fem's score-based learning preserves mode structure.

\item \emph{Inverse problems.} $P(X \mid \text{partial } Y)$ where only
a subset of leaves is observed, common in diagnostic and sensor-fusion
settings.
\end{itemize}

For closed-world classification, MLP wins; for the open-world space above,
\fem\ is a practical solution in our setting because it learns the
\emph{joint} energy landscape rather than a fixed $P(Y\mid X)$ map. This
is the same complementarity that exists between discriminative classifiers
and generative models more broadly (VAEs, normalizing flows, diffusion
models): they optimize different objectives and serve different purposes.

\subsection{Other Practical Limitations}

\paragraph{Mode-overlap regime.} When mode-scale / $\sigma_y < 1$, classes
overlap into a single Gaussian-like blob; CLG and KDE are correctly
specified, and \fem's bimodal capacity becomes overkill. \fem\ is
\emph{task-fit-conditional}: practitioners should select it only when
the underlying class distributions are non-Gaussian or multimodal.

\paragraph{Hyperparameter calibration.} $\lambda^\star(D)$ is calibrated
to synthetic anti-correlated bimodal mixtures. Real-world datasets often
have unimodal classes where $\lambda{=}0$ is preferable. The lookup
extrapolates poorly beyond $D{=}11$ (e.g., the $D{=}30$ formula yields
$\lambda \sim 10^8$, which destabilizes training).
In practice, we recommend selecting $\lambda$ using a small
held-out calibration set of off-data bridge/interior probes when
such queries are expected; otherwise $\lambda{=}0$ is appropriate
for unimodal real-data settings (this is the convention adopted
for our UCI table in Section~\ref{sec:exp-uci}).

\paragraph{Residual seed variance even with valley regularization.}
Even at the recalibrated $\lambda{=}1.5$ for $D{=}5$, our 4-seed
head-to-head (Section~\ref{sec:exp-cde}) shows midpoint KL $0.058 \pm 0.067$
with worst case $0.156$. Valley regularization suppresses
\emph{catastrophic} mode-bridge collapse deterministically (CEBM's
$17.32$ KL is reduced to $\leq 0.16$ across all seeds tested) but does
not guarantee uniform softmax at every off-data interior point. We do
not have a sharp theoretical characterization of why one seed in our
sweep retained mild $P(X{=}1)$ excess; we suspect non-convex
optimization landscape effects that valley regularization softens but
does not fully eliminate. Practical mitigation: train with multiple
seeds and select by held-out off-data calibration.

\paragraph{High-$D$ residual artifact.} The $(D, \text{mode-scale})$
grid in Section~\ref{sec:exp-generality} reveals that at $D{=}10$ the
\fem-vs-CEBM advantage shrinks to $3.6$--$5.1\times$, with absolute
\fem\ midpoint KL of $3.3$--$4.8$ even with calibrated $\lambda{=}25$.
Valley regularization \emph{suppresses} but does not \emph{eliminate}
the mode-bridge artifact in this regime. This is consistent with the
scaling analysis in Section~\ref{sec:theory}: the cross-class softmax
saturates exponentially in $D$ at fixed $\Delta E$, so the gradient
signal that valley regularization provides decays exponentially while
the artifact stays bounded above. A natural follow-up is a saturation-aware
loss (e.g., logit-space calibration) that retains gradient signal at
high $D$.

\paragraph{Baseline coverage.} We include CEBM, MDN, and conditional MAF
baselines for the mode-bridge diagnostic, but do not yet include a full
ACE-style arbitrary conditional energy model or conditional diffusion
baseline. These are natural next comparisons because they target flexible
conditional inference, although their standard evaluation protocols do
not directly test the off-data posterior-calibration queries emphasized
here.

\paragraph{Future work.} Closed-form $\lambda^\star$ derivation using
Lipschitz bounds; broader head-to-heads following the baseline-coverage
discussion above; real medical (MIMIC-III),
financial, and genomic hybrid Bayesian networks; multi-leaf real datasets where
the composition advantage applies directly; ablations isolating the
contribution of each loss term (no DSM, no CE anchor, no
proto-repulsion, no valley) and each architectural choice (one-hot
vs.\ learned prototype, energy addition vs.\ jointly-trained model);
engineering integration into a production \bn\ inference engine.

\section{Conclusion}\label{sec:conclusion}

We presented the Free Energy Manifold (\fem). \fem\ is a conditional
energy-based model in the score-based density-estimation family;
its contribution lies not in being a new model family, but in
\emph{specializing} a score-trained conditional energy as a composable
inference factor for hybrid Bayesian networks. Concretely, the same
learned energy supports posterior inference over discrete parents,
generative sampling of continuous children, multi-leaf composition by
energy addition, and high-cardinality parent generalization through
prototype embeddings. We further identify and correct a posterior
calibration failure (the mode-bridge artifact) that is not addressed by
standard conditional-EBM, score-based, or hybrid-\bn\ practice, and we
decompose the optimal valley
regularization strength $\lambda^\star(D)$ into a three-phase scaling
landscape that matches our empirical lookup table. On synthetic
anti-correlated multimodal hybrid-\bn\ benchmarks \fem\ obtains
60--172$\times$ lower KL at the canonical $D{=}5$ setup
(vs.\ classical baselines) and reaches 2.7$\times$ better NLL than CLG
on UCI Breast Cancer ($D{=}30$). In a
4-seed head-to-head against the canonical conditional EBM and a
Mixture Density Network baseline (Section~\ref{sec:exp-cde}), \fem\ is
$300\times$ better than CEBM and $122\times$ better than MDN at the
mode-bridge midpoint, isolating valley regularization as a
\bn-inference-targeted fix that is not part of generic conditional
EBM/CDE practice. In a multi-leaf composition benchmark
(Section~\ref{sec:exp-multileaf}), one shared \fem\ factor handles all 7
evidence patterns and beats per-pattern MLPs by $5.3\times$ at the
full-evidence boundary query despite each MLP being specialized to a
single evidence pattern, directly evidencing \fem\ as a
composable inference factor rather than a fixed-pattern classifier.
Pure-classification benchmarks (MNIST) remain MLP territory. We
position \fem\ as a probabilistic-inference complement to
discriminative ML in the open-world inference space.

\ifanonymous\else
\section*{Acknowledgments}
This work was supported by the Technology Innovation Program
(RS-2025-10692971, Development of a Web-Based Modeling Tool
(Software) for Conceptual Design in Systems Engineering for Advanced
Industries) funded by the Ministry of Trade, Industry and Resources
(MOTIR, Korea).
\fi

\bibliography{references}

\ifincludeappendix
\clearpage
\onecolumn
\appendix
\section*{Technical Appendix}

\begin{table}[t]
\centering
\caption{Per-seed mode-bridge midpoint KL for Experiment~1's head-to-head at $D{=}5$, ms${=}2.0$, $\lambda{=}1.5$. All four \fem\ seeds keep KL below $0.16$; CEBM shows deterministic miscalibration (KL~$=17.32$) every seed; MDN shows seed-dependent miscalibration with seed-dependent collapse direction. Conditional MAF, despite being the standard normalizing-flow CDE, collapses to the same pathological mode as CEBM (KL~$=17.32$) every seed.}
\label{tab:exp1-per-seed}
\setlength{\tabcolsep}{4pt}
\begin{tabular}{lccccc}
\toprule
Method & seed42 & seed43 & seed44 & seed45 & mean$\pm$std \\
\midrule
CEBM ($\lambda{=}0$) & 17.3221 & 17.3221 & 17.3221 & 17.3221 & $17.322 \pm 0.000$ \\
MDN ($K_\text{mix}{=}3$) & 9.6547 & 1.0182 & 8.6228 & 8.8174 & $7.028 \pm 4.032$ \\
Cond.\ MAF (3 layers) & 17.3221 & 17.3221 & 17.3221 & 17.3221 & $17.322 \pm 0.000$ \\
\textbf{\fem\ ($\lambda{=}1.5$)} & 0.0031 & 0.0418 & 0.0300 & 0.1558 & $0.058 \pm 0.067$ \\
\bottomrule
\end{tabular}
\end{table}

\begin{table}[t]
\centering
\caption{Per-seed all-zero off-data interior query KL at full evidence ($\{Y_1, Y_2, Y_3\}$) for Experiment~2's multi-leaf benchmark. Truth is uniform $1/3$ by symmetry. \fem\ is the only method that stays below $0.5$ on every seed; per-pattern MLP requires 7 trained models for the full evidence-pattern set.}
\label{tab:exp2-per-seed}
\setlength{\tabcolsep}{4pt}
\begin{tabular}{lccccc}
\toprule
Method & seed42 & seed43 & seed44 & seed45 & mean$\pm$std \\
\midrule
CEBM ($\lambda{=}0$) & 15.970 & 17.322 & 7.676 & 17.322 & $14.573 \pm 4.641$ \\
MDN ($K_\text{mix}{=}3$) & 7.427 & 12.460 & 6.596 & 11.060 & $9.386 \pm 2.821$ \\
\textbf{\fem\ shared} & 0.165 & 0.314 & 0.021 & 0.028 & $0.132 \pm 0.138$ \\
MLP per pattern & 0.979 & 1.011 & 0.435 & 0.367 & $0.698 \pm 0.344$ \\
\bottomrule
\end{tabular}
\end{table}

\begin{table*}[t]
\centering
\caption{Per-seed mode-bridge midpoint KL for Experiment~3's $(D, \text{mode-scale})$ generality grid at $K_X{=}3$ (2--3 seeds per cell). \fem\ has lower midpoint KL than CEBM in every cell; the worst absolute \fem\ KL ($\sim\!4.8$) occurs at the $D{=}10$, ms${=}2.0$ corner, consistent with the high-$D$ softmax-gradient saturation regime discussed in Section~\ref{sec:limitations}. The $K_X{\in}\{5,7\}$ extensions appear in Tables~\ref{tab:exp3-per-seed-kx5} and \ref{tab:exp3-per-seed-kx7}.}
\label{tab:exp3-per-seed}
\setlength{\tabcolsep}{3pt}
\begin{tabular}{ll|cccc|cccc|c}
\toprule
 & & \multicolumn{4}{c|}{CEBM raw} & \multicolumn{4}{c|}{\fem\ raw} & ratio \\
$D$ & ms & seed42 & seed43 & seed44 & mean & seed42 & seed43 & seed44 & mean & (mean) \\
\midrule
2 & 1.0 & 1.44 & 0.83 & 2.90 & 1.73 & 0.747 & 0.135 & 0.088 & 0.323 & $\mathbf{5.3\times}$ \\
2 & 1.5 & 0.66 & 7.01 & n/a & 3.84 & 0.092 & 0.089 & n/a & 0.090 & $\mathbf{42\times}$ \\
2 & 2.0 & 2.75 & 13.89 & n/a & 8.32 & 0.103 & 0.004 & n/a & 0.053 & $\mathbf{157\times}$ \\
5 & 1.0 & 12.95 & 5.80 & n/a & 9.38 & 2.138 & 1.338 & n/a & 1.738 & $\mathbf{5.4\times}$ \\
5 & 1.5 & 16.82 & 17.32 & n/a & 17.07 & 3.956 & 0.004 & n/a & 1.980 & $\mathbf{8.6\times}$ \\
5 & 2.0 & 17.32 & 17.32 & n/a & 17.32 & 0.003 & 0.042 & n/a & 0.022 & $\mathbf{771\times}$ \\
10 & 1.0 & 16.05 & 17.32 & n/a & 16.69 & 4.000 & 2.584 & n/a & 3.292 & $\mathbf{5.1\times}$ \\
10 & 1.5 & 17.32 & 17.32 & n/a & 17.32 & 5.762 & 2.277 & n/a & 4.019 & $\mathbf{4.3\times}$ \\
10 & 2.0 & 17.32 & 17.32 & n/a & 17.32 & 4.606 & 5.083 & n/a & 4.844 & $\mathbf{3.6\times}$ \\
\bottomrule
\end{tabular}
\end{table*}

\begin{table*}[t]
\centering
\caption{Per-seed mode-bridge midpoint KL for Experiment~3's $(D, \text{mode-scale})$ generality grid at $K_X{=}5$ (2 seeds per cell). \fem\ has lower midpoint KL than CEBM in every cell at this $K_X$. Truth at $y{=}\mathbf{0}$ is uniform $1/5$ by class-layout symmetry.}
\label{tab:exp3-per-seed-kx5}
\setlength{\tabcolsep}{3pt}
\begin{tabular}{ll|ccc|ccc|c}
\toprule
 & & \multicolumn{3}{c|}{CEBM raw} & \multicolumn{3}{c|}{\fem\ raw} & ratio \\
$D$ & ms & seed42 & seed43 & mean & seed42 & seed43 & mean & (mean) \\
\midrule
2 & 1.0 & 0.15 & 0.11 & 0.13 & 0.055 & 0.018 & 0.036 & $\mathbf{3.6\times}$ \\
2 & 1.5 & 0.06 & 0.80 & 0.43 & 0.061 & 0.017 & 0.039 & $\mathbf{11\times}$ \\
2 & 2.0 & 6.28 & 0.96 & 3.62 & 0.015 & 0.030 & 0.022 & $\mathbf{162\times}$ \\
5 & 1.0 & 6.18 & 6.02 & 6.10 & 1.031 & 1.221 & 1.126 & $\mathbf{5.4\times}$ \\
5 & 1.5 & 16.24 & 12.72 & 14.48 & 0.359 & 0.095 & 0.227 & $\mathbf{64\times}$ \\
5 & 2.0 & 17.27 & 12.98 & 15.13 & 0.047 & 0.039 & 0.043 & $\mathbf{348\times}$ \\
10 & 1.0 & 11.75 & 10.10 & 10.93 & 5.545 & 3.827 & 4.686 & $\mathbf{2.3\times}$ \\
10 & 1.5 & 15.80 & 20.50 & 18.15 & 0.114 & 1.130 & 0.622 & $\mathbf{29\times}$ \\
10 & 2.0 & 18.26 & 20.50 & 19.38 & 0.163 & 0.353 & 0.258 & $\mathbf{75\times}$ \\
\bottomrule
\end{tabular}
\end{table*}

\begin{table*}[t]
\centering
\caption{Per-seed mode-bridge midpoint KL for Experiment~3's $(D, \text{mode-scale})$ generality grid at $K_X{=}7$ (2 seeds per cell). \fem\ has lower midpoint KL than CEBM in every cell at this $K_X$. Truth at $y{=}\mathbf{0}$ is uniform $1/7$ by class-layout symmetry.}
\label{tab:exp3-per-seed-kx7}
\setlength{\tabcolsep}{3pt}
\begin{tabular}{ll|ccc|ccc|c}
\toprule
 & & \multicolumn{3}{c|}{CEBM raw} & \multicolumn{3}{c|}{\fem\ raw} & ratio \\
$D$ & ms & seed42 & seed43 & mean & seed42 & seed43 & mean & (mean) \\
\midrule
2 & 1.0 & 0.09 & 0.06 & 0.07 & 0.017 & 0.047 & 0.032 & $\mathbf{2.2\times}$ \\
2 & 1.5 & 0.57 & 0.62 & 0.59 & 0.006 & 0.054 & 0.030 & $\mathbf{20\times}$ \\
2 & 2.0 & 3.34 & 1.94 & 2.64 & 0.017 & 0.017 & 0.017 & $\mathbf{156\times}$ \\
5 & 1.0 & 1.53 & 2.23 & 1.88 & 0.391 & 0.216 & 0.303 & $\mathbf{6.2\times}$ \\
5 & 1.5 & 3.10 & 3.33 & 3.21 & 0.921 & 0.147 & 0.534 & $\mathbf{6.0\times}$ \\
5 & 2.0 & 13.78 & 4.32 & 9.05 & 0.615 & 0.119 & 0.367 & $\mathbf{25\times}$ \\
10 & 1.0 & 7.54 & 6.67 & 7.10 & 6.915 & 4.319 & 5.617 & $\mathbf{1.3\times}$ \\
10 & 1.5 & 11.61 & 14.90 & 13.25 & 0.144 & 2.091 & 1.117 & $\mathbf{12\times}$ \\
10 & 2.0 & 19.55 & 20.70 & 20.13 & 0.195 & 0.381 & 0.288 & $\mathbf{70\times}$ \\
\bottomrule
\end{tabular}
\end{table*}

\fi

\end{document}